\pdfoutput=1
\documentclass[11pt]{article}

\usepackage{EMNLP2022}
\usepackage{graphicx}
\usepackage{times}
\usepackage{latexsym}
\usepackage{arabtex}
\usepackage[T1]{fontenc}
\usepackage[utf8]{inputenc}

\usepackage{microtype}

\usepackage{inconsolata}
\usepackage{booktabs}

\newcommand\Cstar{CALIMA$_{Star}$}
\newcommand\camelira{Camelira}

\title{Camelira: An Arabic Multi-Dialect Morphological Disambiguator}

\author{Ossama Obeid$^1$, Go Inoue$^{1,2}$, Nizar Habash$^1$ \\
  $^1$Computational Approaches to Modeling Language (CAMeL) Lab \\
  New York University Abu Dhabi \\
  $^2$Mohamed bin Zayed University of Artificial Intelligence \\
  {\tt\{oobeid,nizar.habash\}@nyu.edu} \\
  {\tt go.inoue@mbzuai.ac.ae} \\}

\begin{document}
\maketitle
\begin{abstract} % Go
We present \camelira, a web-based Arabic multi-dialect morphological disambiguation tool that covers four major variants of Arabic: Modern Standard Arabic, Egyptian, Gulf, and Levantine.
{\camelira} offers a user-friendly web interface that allows researchers and language learners to explore various linguistic information, such as part-of-speech, morphological features, and lemmas.
Our system also provides an option to automatically choose an appropriate dialect-specific disambiguator based on the prediction of a dialect identification component. {\camelira} is publicly accessible at  \url{http://camelira.camel-lab.com}.
%$^,$\footnote{Video: \url{http://camelira-vid.camel-lab.com}.}
%{\camelira} leverages the state-of-the-art morphological disambiguation models and dialect identification system.
%Our demo will be publicly available at \url{http://camelira.camel-lab.com}.
%{\camelira} takes a sentence as input and provides an automatically disambiguated reading for each word, as well as its alternative readings.
\end{list}
\end{abstract}

\setarab
\novocalize

\section{Introduction} % Go
\label{sec:intro}
The last two decades have witnessed remarkable progress in Natural Language Processing (NLP) for Arabic and its dialects despite many challenges such as its diglossic nature, morphological complexity, and orthographic ambiguity~\cite{darwish2021panoramic}.
These efforts have led to many practical applications for various NLP tasks including tokenization, part-of-speech (POS) tagging, morphological disambiguation, named entity recognition, dialect identification (DID), and sentiment analysis~\cite[inter alia]{Pasha:2014:madamira,Abdelali:2016:farasa,obeid-etal-2019-adida,abdul-mageed-etal-2020-aranet}.

Tools for core technologies like POS tagging and morphological disambiguation are primary examples of such successful applications, e.g., MADAMIRA~\cite{Pasha:2014:madamira}, Farasa~\cite{Abdelali:2016:farasa},  UDPipe~\cite{straka-etal-2016-udpipe}, and Stanza~\cite{qi-etal-2020-stanza}.
However, there are still gaps to be filled in terms of coverage and usability.
For example, these systems only support Modern Standard Arabic (MSA) and Egyptian Arabic, but not other widely spoken dialects such as Gulf and Levantine.
In addition, these web interfaces only present the top prediction, although the alternative readings could provide valuable information for analyzing the models' behavior.
In contrast, morphological analyzers such as ElixirFM~\cite{Smrz:2007:elixirfm}, {\Cstar}~\cite{taji-etal-2018-arabic}, CALIMA Egyptian~\cite{Habash:2012:morphological} show all the different readings for a given word out of context but without disambiguated analyses in context.
These tools assume that users already know the input DID; however, this is not necessarily the case for second language learners.

To address these limitations, we present \camelira,\footnote{\url{http://camelira.camel-lab.com}}{$^,$}\footnote{{\camelira} is named after CAMeL Tools~\cite{Obeid:2020:Cameltools}, and in homage to MADAMIRA \cite{Pasha:2014:madamira}.} a web interface for Arabic multi-dialect morphological disambiguation that covers four major variants of Arabic: MSA, Egyptian, Gulf, and Levantine.
Our system takes an input sentence and provides  automatically disambiguated readings for each word in context, as well as its alternative out-of-context readings.
We also showcase the integration of a state-of-the-art morphological disambiguator~\cite{inoue-etal-2022-morphosyntactic} with the highest performing fine-grained Arabic DID system~\cite{Salameh:2018:fine-grained} on the MADAR DID shared task \cite{bouamor-etal-2019-madar}.
{\camelira} provides an option to automatically choose a dialect-specific disambiguator based on the prediction of the DID component.
To the best of our knowledge, our work is the first to demonstrate an integrated web application that leverages both Arabic morphological disambiguation and DID systems.

Our contributions are as follows:
(a) We present a user-friendly web interface that allows researchers and language learners to explore the detailed linguistic analysis of a given Arabic sentence.
(b) We include three major Arabic dialects (Egyptian, Gulf, and Levantine) in addition to MSA, to make our tool more accessible to a wider audience.
(c) We integrate DID to automatically select the appropriate disambiguator; a feature that helps users with limited knowledge of Arabic dialects.

%Our demo will be publicly available at \url{http://camelira.camel-lab.com}.

\section{Arabic Linguistic Facts}
\label{sec:ling}
The Arabic language poses a number of challenges for NLP~\cite{Habash:2010:introduction}.
We highlight three aspects that are most relevant to multi-dialectal morphological modeling: dialectal variations, morphological richness, and orthographic ambiguity.

First, Arabic is characterized with diglossia and its large number of dialects~\cite{Ferguson:1959:diglossia,Holes:2004:modern}.
MSA is the shared standard variant used in official contexts, while the dialects are the varieties of daily use.
MSA and the dialects vary among themselves in different aspects, such as lexicons, morphology, and syntax.
Second, Arabic is a morphologically rich and complex language.
It employs a combination of templatic, affixational, and cliticization morphological operations to represent numerous grammatical features such as gender, number, person, case, state, mood, aspect, and voice, in addition to a number of attachable pronominal, preposition, and determiner clitics.
Third, Arabic is orthographically highly ambiguous.
This is due to its orthographic conventions where diacritical marks are often omitted, leading to a high degree of ambiguity.
For example, MSA can have 12 different morphological analyses per word on average~\cite{Pasha:2014:madamira}.

\section{Related Work}
\label{sec:related}
\paragraph{Morphological Analysis and Disambiguation}
Morphological analysis is the task of producing a complete list of readings (analyses) for a given word out of context.
Morphological analysis has a wide range of applications, including treebank annotation~\cite{Maamouri:2003:patb-1,Maamouri:2011:patb-2,Maamouri:2009:patb-3} and improving morphological modeling~\cite{Habash:2005:morphological,Inoue:2017:joint,Zalmout:2017:dont,khalifa-etal-2020-morphological}.
Over the past two decades, there have been numerous efforts in building morphological analyzers for Arabic, e.g.
BAMA~\cite{Buckwalter:2002:buckwalter},
MAGEAD~\cite{Habash:2006:magead,Altantawy:2010:morphological}, 
ALMORGEANA~\cite{Habash:2007:arabic-representation},
ElixirFM~\cite{Smrz:2007:elixirfm},
SAMA~\cite{Graff:2009:standard},
CALIMA Egyptian \cite{Habash:2012:morphological},
CALIMA Gulf \cite{Khalifa:2017:morphological},
AlKhalil Morpho Sys~\cite{Boudlal:2010:alkhalil,Boudchiche:2017:alkhalil}
and \Cstar~\cite{taji-etal-2018-arabic}.
Among these efforts,
ElixirFM\footnote{\url{http://quest.ms.mff.cuni.cz/elixir}} and \Cstar\footnote{\url{http://calimastar.camel-lab.com/}} provide easy-to-use web interfaces, allowing the user to explore all the possible morphological analyses for a given word.
In addition to these rule-based approaches, \newcite{Eskander:2016:creating} used a corpus-based paradigm completion technique \cite{Eskander:2013:automatic-extraction} to develop a morphological analyzer for Levantine; and \cite{khalifa-etal-2020-morphological} used the same technique to develop a morphological analyzer for Gulf.

Morphological disambiguation is the subsequent process of identifying the correct analysis in context from the list of different analyses produced by a morphological analyzer.
Examples of this in Arabic start with  MADA~\cite{Habash:2005:morphological} and many following efforts \cite{Pasha:2014:madamira,Khalifa:2016:yamama,Zalmout:2017:dont,
%Zalmout:2018:noise-robust,zalmout-habash-2019-adversarial,
zalmout-habash-2020-joint,khalifa-etal-2020-morphological,inoue-etal-2022-morphosyntactic}, where they rank the analyses based on the predictions of morphological taggers.
While these models have achieved significant improvement over time, only MADAMIRA~\cite{Pasha:2014:madamira} offers a web interface\footnote{\url{http://madamira.camel-lab.com/}} that's accessible to a general audience.
In this work, we present a user-friendly web interface for  state-of-the-art morphological disambiguation models to make these recent advances more accessible to a wider audience, such as linguists and language learners.
Our interface also provides all the alternative readings of each input word with the associated prediction scores, allowing researchers to investigate the model's behavior.

\begin{figure*}[t]
\centering
\frame{\includegraphics[width=1.9\columnwidth]{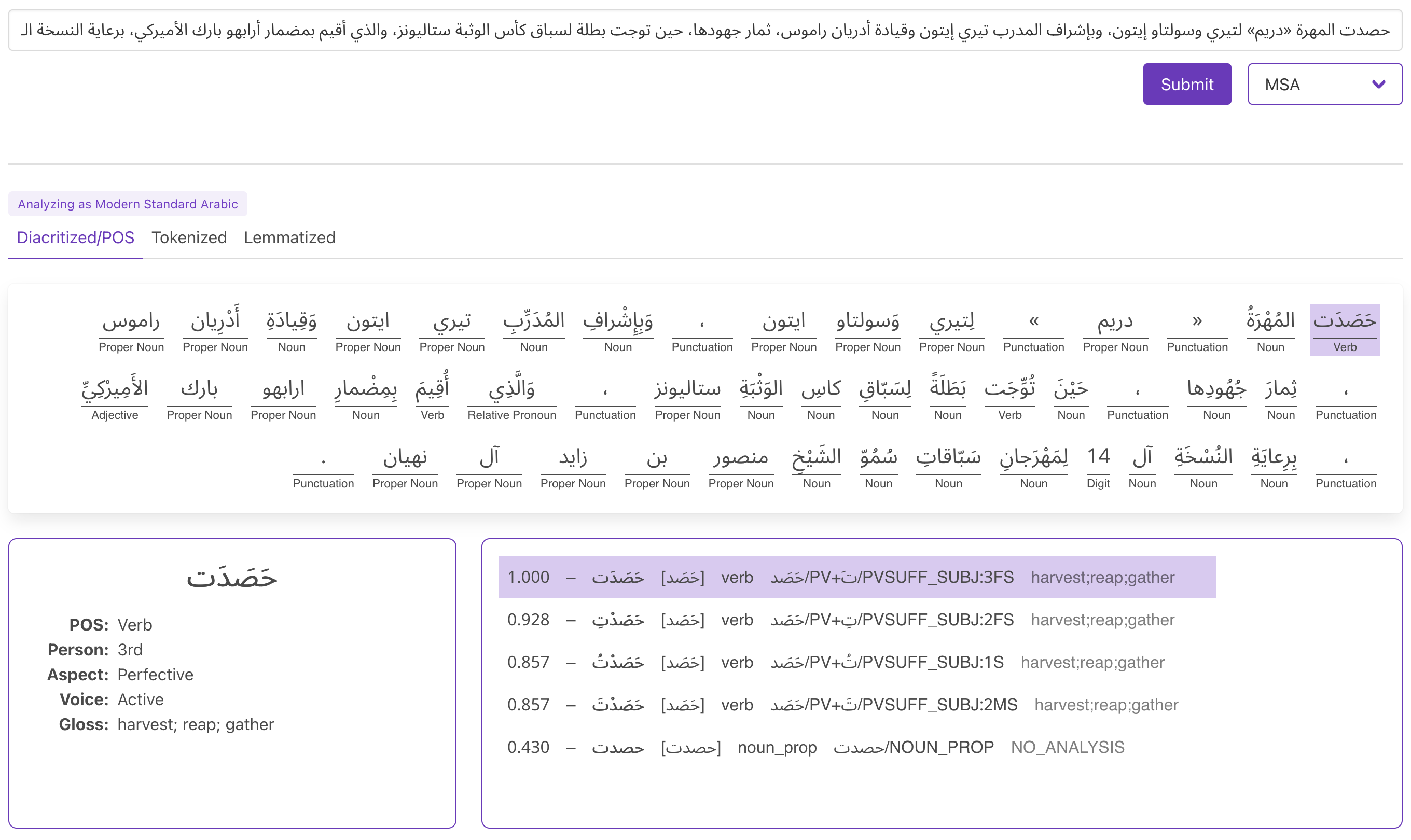}}
\caption{
The {\camelira} interface with an MSA example sentence celebrating the winning of a racehorse named ``Dream.''
In this example, the automatically diacritized forms of the words are presented together with their POS.
The first word (on the right), which is highlighted, is selected by the user.
The two lower boxes show all the possible out-of-context analyses (on the right) and the detailed features and gloss for the top in-context analysis (on the left).
}
%\vspace{6pt}
\label{fig:msapos}
\end{figure*}

\begin{figure*}[t]
\centering
\frame{\includegraphics[width=1.9\columnwidth]{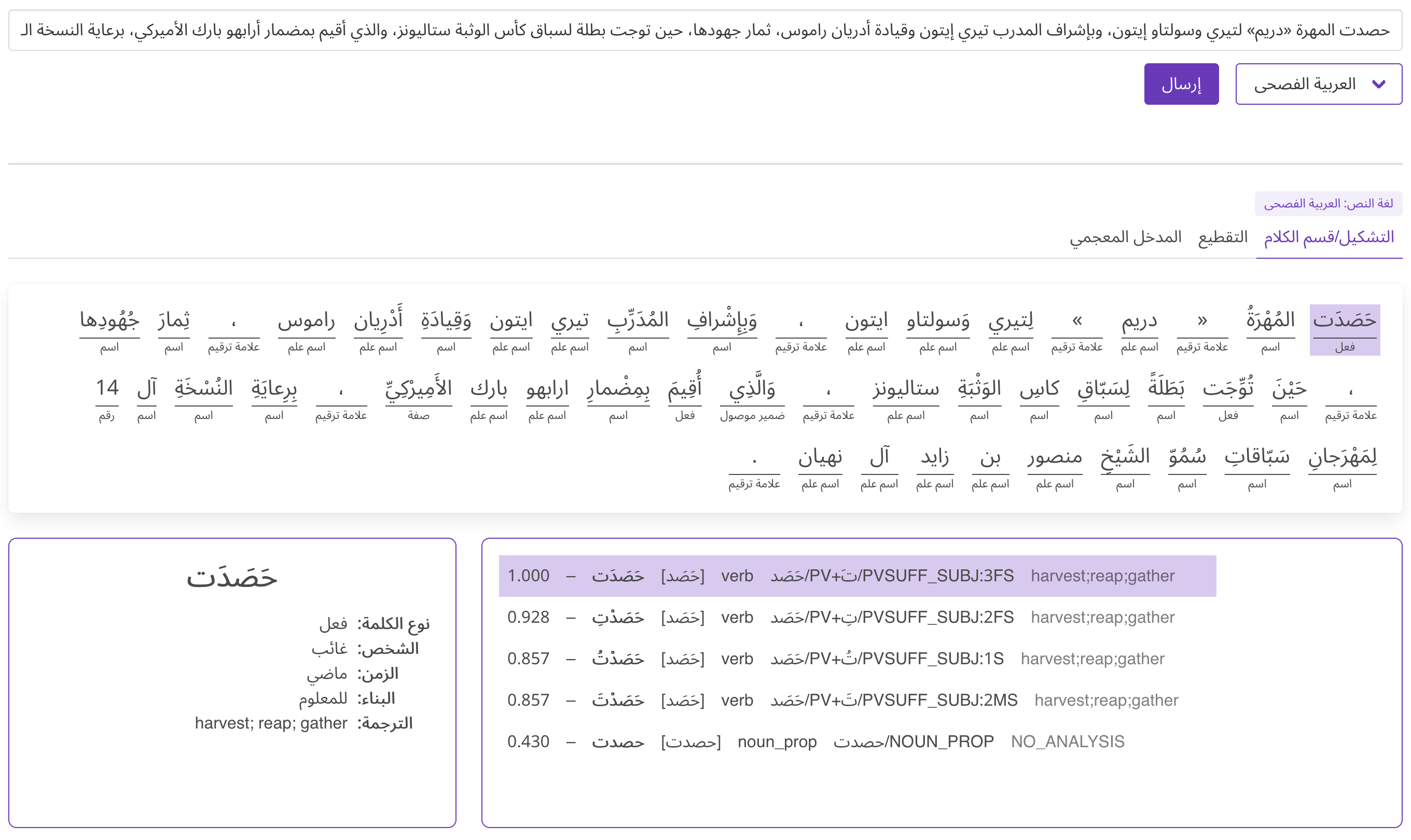}}
\caption{
The {\camelira} interface presenting the same example in Figure~\ref{fig:msapos} using the Arabic user interface.
}
%\vspace{6pt}
\label{fig:msaposar}
\end{figure*}

\paragraph{Dialect Identification}
Dialect identification (DID) is the task of automatically identifying the language variety of a given text.
DID for Arabic and its variants has attracted increasing attention in recent years.
A number of shared tasks have been organized, including VarDial~\cite{malmasi-etal-2016-discriminating,zampieri-etal-2017-findings,zampieri-etal-2018-language}, MADAR~\cite{bouamor-etal-2019-madar}, and NADI~\cite{abdul-mageed-etal-2020-nadi,abdul-mageed-etal-2021-nadi,mageed:2022:nadi}, along with  continuous efforts in dataset creation~\cite[inter alia]{Zaidan:2011:arabic,Mubarak:2014:using,Zaghouani:2018:araptweet,baimukan-bouamor-habash:2022:LREC}.
These evaluation campaigns have led to the development of practical applications, such as ADIDA\footnote{\url{http://adida.camel-lab.com/}}~\cite{obeid-etal-2019-adida}, a web interface for fine-grained Arabic DID based on the highest performing system in the MADAR shared task~\cite{Salameh:2018:fine-grained}.
In this work, we employ one of the DID systems described by \newcite{Salameh:2018:fine-grained};\footnote{We use regional level classification instead of fine-grained city-level classification because the morphological analyzers are designed at the regional level.} however, we differ from their work in that we combine DID with multi-dialect morphological disambiguation to allow users to easily select an appropriate dialect-specific Arabic disambiguator based on the DID prediction.
%In this work, we employ the same DID system\footnote{We use regional level classification instead of fine-grained city-level classification because the morphological analyzers are designed at the regional level.} implemented in CAMeL Tools~\cite{Obeid:2020:Cameltools}; however, we differ from their work in that we combine DID with multi-dialect morphological disambiguation to allow users to easily select an appropriate dialect-specific Arabic disambiguator based on the DID prediction.

\section{System Design and Implementation} % Ossama
\label{sec:design}
\subsection{Design Considerations}
We want an easy-to-use one-stop online-accessible user interface that supports the analysis of Arabic sentences from different dialects, and with access to under-the-hood decisions about disambiguation.
To that end, we are inspired by three web interfaces: MADAMIRA \cite{Pasha:2014:madamira} for in-context disambiguation, {\Cstar} \cite{Taji:2018:arabic} for out-of-context analysis, and ADIDA \cite{obeid-etal-2019-adida} for dialect identification.  
Furthermore, we would like the web interface to have a responsive design with streamlined user experiences across a range of devices from mobile to desktops.

%We want an easy-to-use web interface that provides similar functionality to MADAMIRA while providing a more streamlined user experience and a wide compatibility across display devices.
%We also want to provide users with alternative analyses including the top analysis.

\subsection{Implementation}
\paragraph{Back-end}
The back-end is implemented in Python using \texttt{Flask}\footnote{\url{https://flask.palletsprojects.com/}} to serve a REST API.
We implemented the MODEL-6 DID system described by 
\newcite{Salameh:2018:fine-grained} for automatic dialect identification and the morphological disambiguation system described by \newcite{inoue-etal-2022-morphosyntactic}.
The implementation of the morphological disambiguator was provided by the CAMeL Tools\footnote{\url{https://github.com/CAMeL-Lab/camel_tools}} Python API \cite{Obeid:2020:Cameltools}.
We plan to add our MODEL-6 implementation to CAMeL Tools. 

For morphological disambiguation, we use the \textit{unfactored} model with a morphological analyzer for all variants.
We chose the unfactored models because they are faster than the factored models and only slightly lower in performance.
Table \ref{table:morph} shows the performance accuracy of Camelira's morphological disambiguation models.
We report numbers on DEV as presented in \newcite{inoue-etal-2022-morphosyntactic}.

%For DID, we evaluate our MODEL-6 implementation using the DEV and TEST splits of the dataset used by \newcite{Salameh:2018:fine-grained} where it was trained using the TRAIN split. 
%
For DID, we train our MODEL-6 using the TRAIN split and evaluate using the DEV and TEST splits following \newcite{Salameh:2018:fine-grained}.
Table \ref{table:did} compares the performance of our implementation with that of \newcite{Salameh:2018:fine-grained}.
Our results are slightly lower due to implementation differences.

% https://docs.google.com/spreadsheets/d/16wCfWnQynNn8JrH-kW0lMF5GZiMvG3wPBaiXjXJlStY/edit#gid=1458272531

\begin{table}[ht]
\centering
\begin{tabular}{lcc}
\toprule
     & \textbf{ALL TAGS} &  \textbf{POS}\\
\hline
\textbf{MSA}         & 95.9                                  & 98.7                             \\
\textbf{EGY}         & 90.5                                  & 94.0                             \\
\textbf{GLF}         & 93.8                                  & 96.6                             \\
\textbf{LEV}         & 85.5                                  & 92.7                               \\
\bottomrule
\end{tabular}
\caption{
Accuracy of Camelira's morphological disambiguation models based on \newcite{inoue-etal-2022-morphosyntactic}'s unfactored+Morph models.
%, reported in \newcite{inoue-etal-2022-morphosyntactic}.
\textbf{ALL TAGS} is the accuracy of the combined morphosyntactic features.
}
\label{table:morph}
\end{table}

\begin{table}[ht]
\centering
\begin{tabular}{lcc}
\toprule
     & \textbf{DEV} &  \textbf{TEST}\\
\hline
\textbf{Camelira}         & 92.8                             & 93.5                                  \\
%\newcite{Salameh:2018:fine-grained}         & 93.1     
\textbf{Salameh et al.}         & 93.1                             & 93.6                                  \\
\bottomrule
\end{tabular}

\caption{Accuracy of Camelira's implementation of the MODEL-6 DID model compared with  \newcite{Salameh:2018:fine-grained}'s implementation of the same model.
\label{table:did}
}
\end{table}

\paragraph{Front-end}
The front-end was implemented using \texttt{Vue.js}\footnote{\url{https://vuejs.org/}} for model view control and \texttt{Bulma}\footnote{\url{https://bulma.io/}} for styling and creating a responsive design that works well across devices. 

\begin{figure*}[t]
\centering
\frame{\includegraphics[width=1.9\columnwidth]{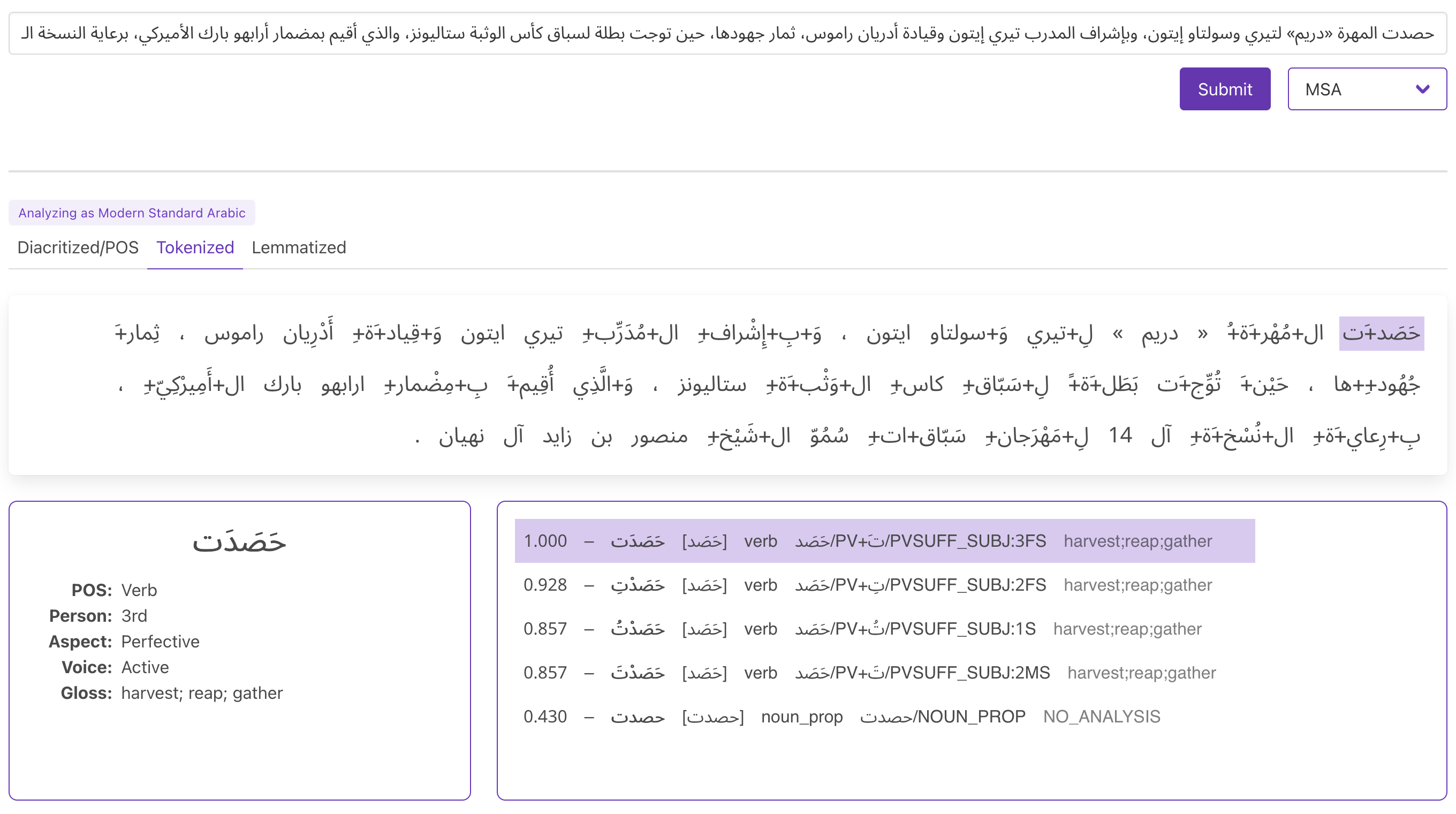}}
\caption{
The {\camelira} interface with an MSA example sentence and ``Tokenized'' display tab.
This is an exact replica of the input and output choices as in Figure~\ref{fig:msapos} except that the word forms are presented in full tokenization.
}
\vspace{3pt}
\label{fig:msatok}
\end{figure*}

\subsection{The {\camelira} Interface}

The {\camelira} interface is divided into three main areas, the Input Area, Text Output Area, and Morphological Analysis Area.
Figure~\ref{fig:msapos} shows an example of a disambiguated MSA sentence in the {\camelira} web interface. We also provide the option of viewing the interface in Arabic as seen in Figure~\ref{fig:msaposar}.

\paragraph{Input Area}
At first, only the Input Area is displayed which provides users with an input box where they can enter the sentence they wish to disambiguate. 
Users are also presented with a drop-down menu where they can select whether to disambiguate the input sentence as a particular dialect (MSA, Egyptian, Gulf, or Levantine) or to have the dialect be automatically selected.

\paragraph{Text Output Area}
Once the submit button is clicked and the sentence has been disambiguated, the Text Output Area is displayed.
First, the dialect indicator displays which dialect was used to analyze the provided input.
Then, an output box displays the disambiguated sentence in three different views:
(a) the \textbf{Diacritized/POS} view which displays the diacritized text (if supported by the selected dialect's resources) along with the POS tag of each word,
(b) the \textbf{Tokenized} view which displays each disambiguated word in its tokenized form where tokens are delimited by a `+' character, and
(c) the \textbf{Lemmatized} view where each word is displayed in its lemmatized form.
Figure~\ref{fig:msatok} is the same as Figure~\ref{fig:msapos} except that the text output is in Tokenized mode.

\paragraph{Morphological Analysis Area}
Below the Text Output Area, the Morphological Analysis Area consists of the Analysis List box (on the right), which displays all analyses of a given word sorted by their disambiguation ranking order, and the Analysis Viewer box (on the left), which displays a selected analysis in an easy-to-read form with more morphological feature details.
The analysis list displays the disambiguation score of each analysis as well as the values for a reduced set of features. 

Clicking on a word in the Text Output Area selects that word, displaying its analyses in the analysis list and analysis viewer boxes.
Clicking on an analysis in the Analysis List will display its user-friendly form in the Analysis Viewer.
By default, the top analysis is selected.

\paragraph{Dialect Identification and Morphological Disambiguation}

\begin{figure}[t]
\centering
\frame{\includegraphics[width=0.95\columnwidth]{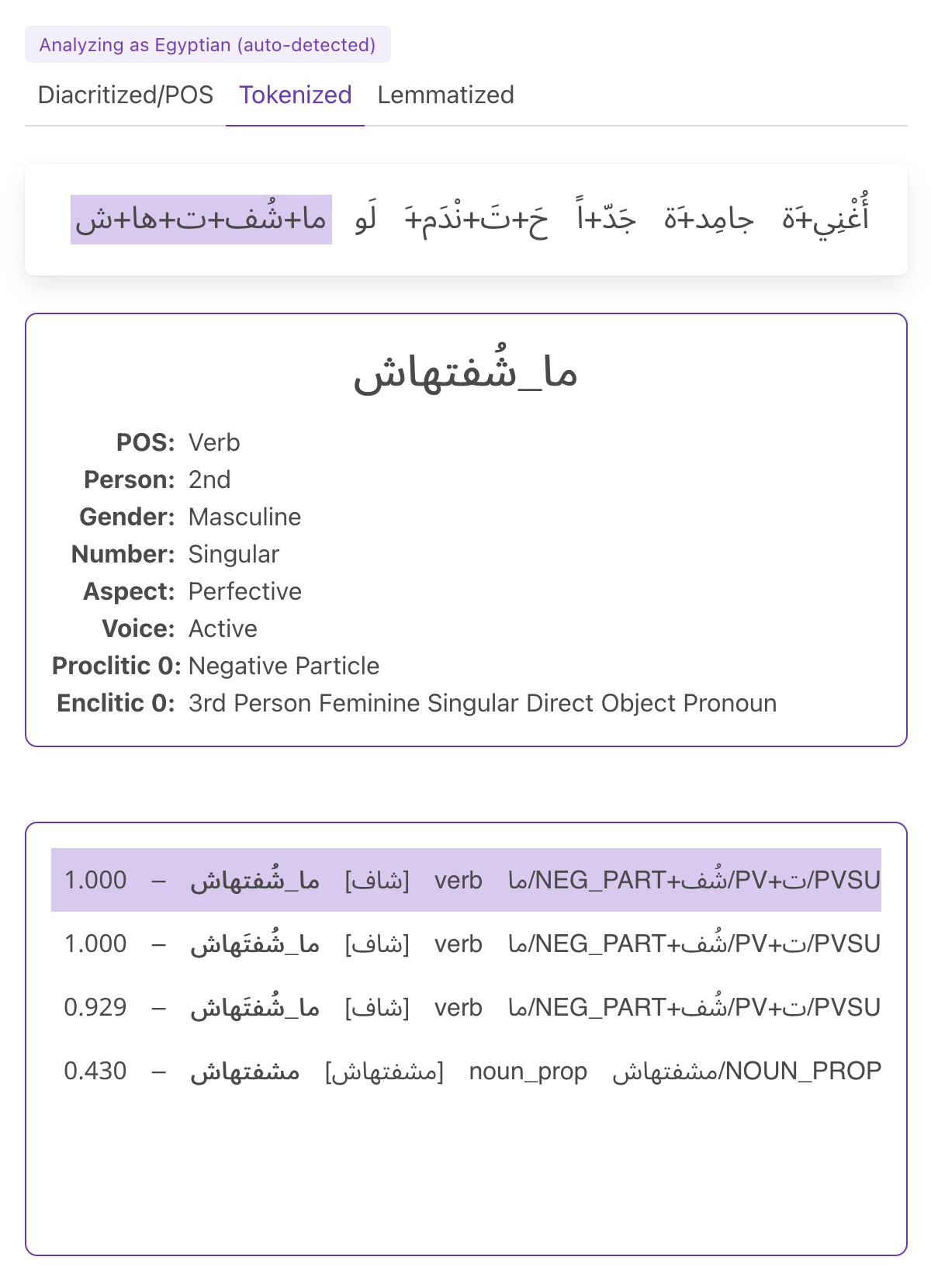}}
\caption{
The {\camelira} interface with an Egyptian example sentence:
\emph{"A very cool song [video clip], you'll regret it if you don't watch it."}
In this example, the input text is automatically correctly detected as Egyptian.
%Translation: \emph{"Very cool songs, you'll regret it if you don't listen to them."}
}
\label{fig:egy}
\end{figure}

\begin{figure}[t]
\centering
\frame{\includegraphics[width=0.95\columnwidth]{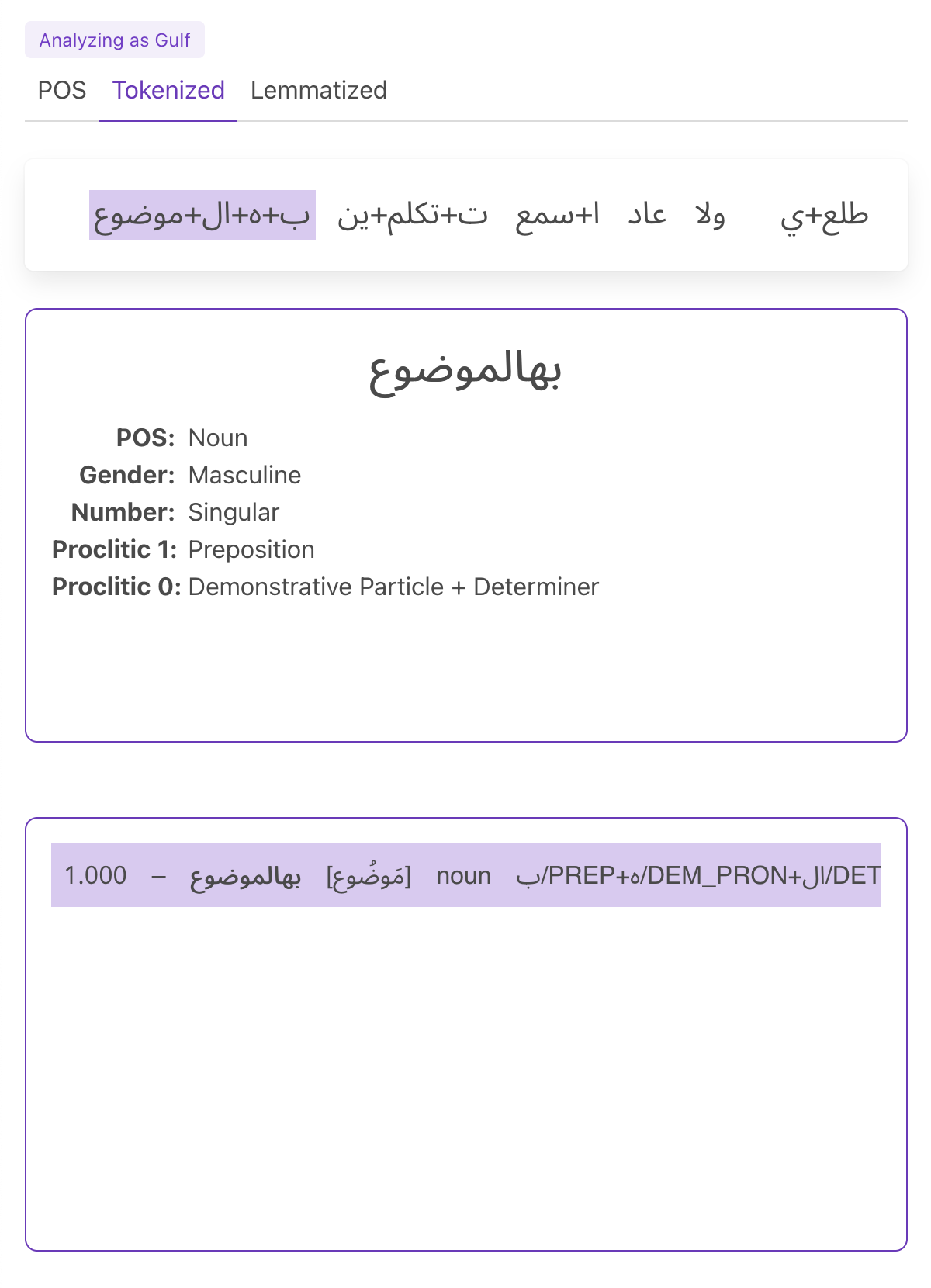}}
\caption{
The {\camelira} interface with a Gulf example sentence: 
\emph{"Go up to your room, I don't want to hear you talking about this subject again."}
In this example, the user specified the input dialect as Gulf.
%Translation: \emph{"Go up to your room, I don't want to hear you talking about this subject again."}
}
\label{fig:glf}
\end{figure}

Figures~\ref{fig:egy}~and~\ref{fig:glf} present Egyptian and Gulf Arabic examples, respectively.
Both are presented in a mobile setting to demonstrate our responsive design.

In the case of Figure~\ref{fig:egy}, the user selected Auto-Detect for dialect identification.
In the Gulf example, the user selected Gulf Arabic directly.
Note that the Gulf Arabic does not show diacritizations since its training data did not include diacritized forms \cite{khalifa-etal-2020-morphological}.

\section{Conclusion and Future Work} % Go
We presented \camelira, a user-friendly web interface for Arabic multi-dialect morphological disambiguation that covers four major variants of Arabic.
The system takes a sentence as input and provides an automatically disambiguated reading for each word, as well as its alternative readings, allowing users to explore various linguistic information, such as part-of-speech, morphological features, and lemmas.
{\camelira} also provides an option to automatically choose an appropriate dialect-specific disambiguator based on the prediction of its dialect identification component.
% to the best of our knowledge, ....

In the future, we plan to extend our disambiguation system to cover other Arabic dialects such as Maghrebi and Yemeni Arabic.
We also plan to continue to update the system using future improvements in terms of efficiency and accuracy in CAMeL Tools~\cite{Obeid:2020:Cameltools}.

\section*{Limitations and Ethical Considerations} % Go
We acknowledge that our system is currently limited to specific variants of Arabic and it can produce erroneous predictions especially on different dialects, genres, and styles that are not covered in the current system's training data.
We also acknowledge that our work on core and generic NLP technologies can be used as part of the pipeline of other systems with malicious intents.

\section*{Acknowledgements}
Some of this work was carried out on the High Performance Computing resources at New York University Abu
Dhabi. 
We thank Salam Khalifa and Bashar Alhafni for their insightful comments and helpful discussions.
We also thank anonymous reviewers for their helpful comments.

\bibliography{anthology,camel-bib-v2,extra}
\bibliographystyle{acl_natbib}

% \appendix

% \section{Example Appendix}
% \label{sec:appendix}

% This is a section in the appendix.

\end{document}